\documentclass[10pt,twocolumn]{article} 
\usepackage{simpleConference}
\usepackage{times}
\usepackage{graphicx}
\usepackage{url,hyperref}
\usepackage{colortbl}
\usepackage[dvipsnames]{xcolor}
\usepackage{comment}
\usepackage{amsmath,amssymb} 
\usepackage{color}
\usepackage{cite}
\usepackage{mathtools}

\usepackage{kotex}
\usepackage{bm}
\usepackage{algorithm}
\usepackage{multirow}
\usepackage{algpseudocode}
\usepackage{lipsum, booktabs}
\newcolumntype{C}[1]{>{\centering\arraybackslash}m{#1}}
\usepackage{tabu}
\usepackage{booktabs}
\usepackage{kotex}
\usepackage{caption}
\usepackage[symbol]{footmisc}

\title{Unified Negative Pair Generation toward Well-discriminative Feature Space for Face Recognition}

\author{Junuk Jung, Seonhoon Lee, Heung-Seon Oh\footnotemark[1], Yongjun Park, Joochan Park, Sungbin Son  \\ \\
  School of Computer Science and Engineering \\
  Korea University of Technology and Education (KOREATECH) \\ \\
\texttt{\{rnans33, karma1002, ohhs, qkr2938, green669, sbson0621\}@koreatech.ac.kr}
}

\begin{document}
\twocolumn[{%
\renewcommand\twocolumn[1][]{#1}%
\maketitle
\begin{center}
    \centering
    \captionsetup{type=figure}
    \includegraphics[width=0.9\textwidth]{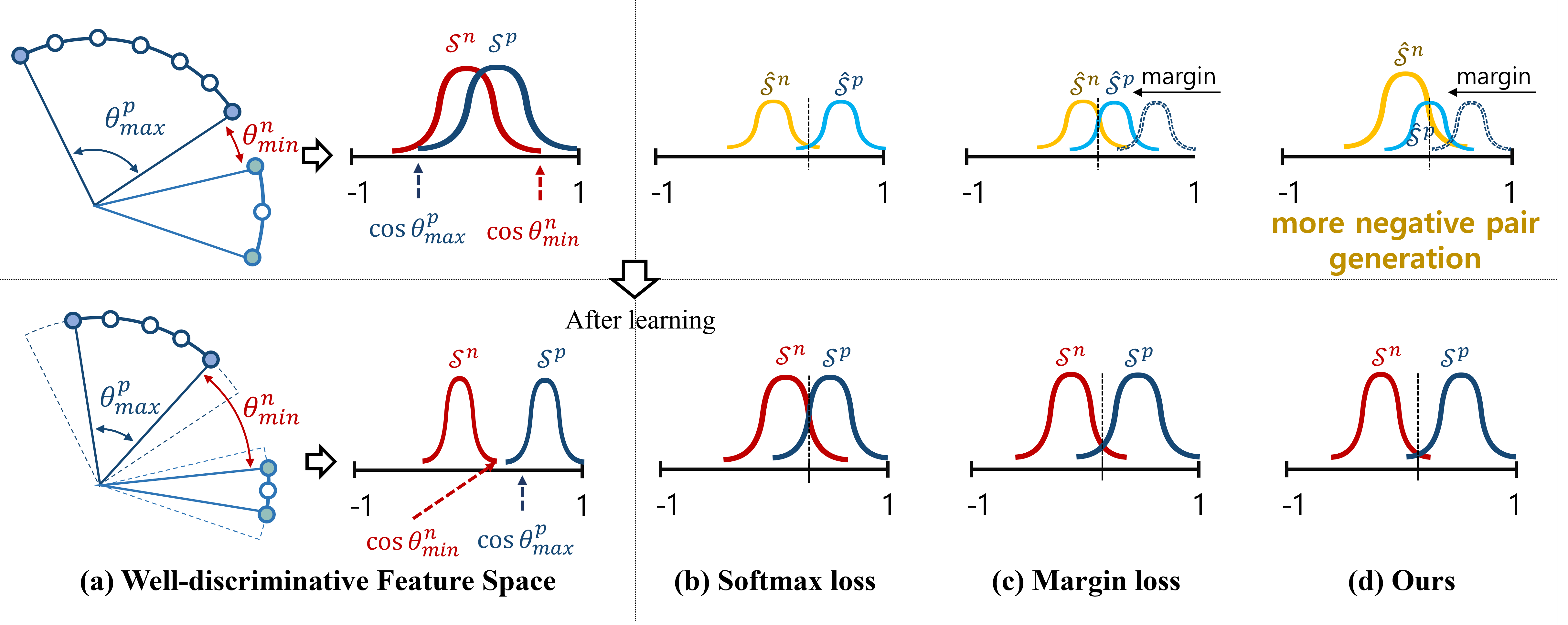}
    \captionof{figure}{Similarity distributions viewed from pair generation perspective for approximating WDFS. The bottom line presents similarity distributions in feature space after sufficiently learning in their own ways with the top line. $\mathcal{S}^p$ and  $\mathcal{S}^n$ are positive and negative similarity sets and $\mathcal{\hat{S}}^p$ and  $\mathcal{\hat{S}}^n$ are subsets of $\mathcal{S}^p$ and $\mathcal{S}^n$, respectively. (a) The ideal similarity sets satisfying $\inf{\mathcal{S}^p} > \sup{\mathcal{S}^n}$ after learning with $\mathcal{S}^p$ and $\mathcal{S}^n$. $\theta^{p}_{max}$ and $\theta^{n}_{min}$ are the max and min angles among positive and negative pairs.  (b) Using a vanilla loss, no overlap between $\mathcal{\hat{S}}^p$ and $\mathcal{\hat{S}}^n$ results in an overlap between $\mathcal{S}^p$ and $\mathcal{S}^n$. (c) Using a marginal loss, an overlap between $\mathcal{\hat{S}}^p$ and $\mathcal{\hat{S}}^n$ by shifting $\mathcal{\hat{S}}^p$ reduces an overlap after learning. (d) Using more negative pairs, an overlap between $\mathcal{\hat{S}}^p$ and $\mathcal{\hat{S}}^n$ by shifting $\mathcal{\hat{S}}^p$ and enlarging $\mathcal{\hat{S}}^n$ significantly reduces the overlap after learning.}
    \label{fig:intro}
\end{center}%
}]

\footnotetext[1]{corresponding author}

\begin{abstract}
The goal of face recognition (FR) can be viewed as a pair similarity optimization problem, maximizing a similarity set $\mathcal{S}^p$ over positive pairs, while minimizing similarity set $\mathcal{S}^n$ over negative pairs. Ideally, it is expected that FR models form a well-discriminative feature space (WDFS) that satisfies $\inf{\mathcal{S}^p} > \sup{\mathcal{S}^n}$. With regard to WDFS, the existing deep feature learning paradigms (i.e., metric and classification losses) can be expressed as a unified perspective on different pair generation (PG) strategies. Unfortunately, in the metric loss (ML), it is infeasible to generate negative pairs taking all classes into account in each iteration because of the limited mini-batch size. In contrast, in classification loss (CL), it is difficult to generate extremely hard negative pairs owing to the convergence of the class weight vectors to their center. This leads to a mismatch between the two similarity distributions of the sampled pairs and all negative pairs. Thus, this paper proposes a unified negative pair generation (UNPG) by combining two PG strategies (i.e., MLPG and CLPG) from a unified perspective to alleviate the mismatch. UNPG introduces useful information about negative pairs using MLPG to overcome the CLPG deficiency. Moreover, it includes filtering the similarities of noisy negative pairs to guarantee reliable convergence and improved performance. Exhaustive experiments show the superiority of UNPG by achieving state-of-the-art performance across recent loss functions on public benchmark datasets.  Our code and pretrained models are publicly available\footnotemark[2].
\end{abstract}

\footnotetext[2]{https://github.com/Jung-Jun-Uk/UNPG.git}

\section{Introduction}
The goal of face recognition (FR) can be viewed as a pair similarity optimization problem that maximizes a similarity set $\mathcal{S}^p$ over positive pairs while minimizing a similarity set $\mathcal{S}^n$ over negative pairs. Regardless of FR tasks such as face verification (1:1) and face identification (1:N), it is expected that FR models form a well-discriminative feature space (WDFS) that satisfies $\inf{\mathcal{S}^p} > \sup{\mathcal{S}^n}$ as shown in Fig. \ref{fig:intro} (a). To this end, previous research advances pair similarity optimization\cite{sun2020circle, wang2019multi, yu2019deep, liu2017sphereface, deng2019arcface, wang2018cosface, meng2021magface} by enhancing intra-class compactness and inter-class dispersion.

In deep feature learning paradigms for pair similarity optimization, loss functions in FR can be categorized based on two approaches: metric loss (ML; e.g., triplet loss\cite{schroff2015facenet, hoffer2015deep} and N-pair loss\cite{sohn2016improved}) and classification loss (CL; e.g., softmax loss\cite{cao2018vggface2,deepface-recognition,taigman2014deepface}). The former directly performs the optimization with a pair of deep feature vectors using a pair-wise label whereas the latter performs indirectly with a pair of deep feature and class weight vectors using a class-level label. Recently, in circle loss\cite{sun2020circle}, two different approaches were expressed as a unified loss function since their intent and behaviors pursued the same objective of maximizing a similarity set $\mathcal{S}^p$ over positive pairs, while minimizing it over negative pairs. We decomposed the unified loss function into pair generation (PG) and similarity computation (SC) without loss of generality. While SC focuses on computing the similarity between two samples in a pair, PG focuses on generating a pair using vectors of deep features or class weights. In the unified loss function, the only difference between ML and CL is PG, because various methods in SC can be applied to both ML and CL in the same manner. Consequently, the core of FR research from a unified perspective is the generation of informative pairs, i.e., PG. This is crucial because only a limited number of pairs are trainable in each iteration owing to the large computational costs incurred. Based on the assumption that pairs sampled from mini-batches can represent the entire feature space, the existing methods decrease the loss to a pair as it approaches WDFS, whereas they increase the loss in the opposite case.

We observed the reason why FR models trained sufficiently fail to approach WDFS. This stems from the mismatch of similarity distributions between the sampled pairs and all pairs. Fig. \ref{fig:intro} (b) shows an example of two similarity sets $\mathcal{\hat{S}}^p$ and $\mathcal{\hat{S}}^n$ of positive and negative pairs, respectively, sampled from mini-batches. A FR model is rarely trainable with $\mathcal{\hat{S}}^p$ and $\mathcal{\hat{S}}^n$ because they are well-separated with almost no overlap. To deal with this problem, a line of research\cite{sun2020circle, wang2019multi, yu2019deep, liu2017sphereface, deng2019arcface, wang2018cosface, meng2021magface} devises marginal loss functions to reduce the gap by shifting $\mathcal{\hat{S}}^p$, as shown in Fig. \ref{fig:intro} (c). In general, marginal CL functions show better performance than ML functions on large-scale datasets\cite{guo2016ms}. However, there still exists a mismatch between the sampled pairs and all pairs because it is difficult to obtain too-hard negative pairs.

This paper proposes unified negative pair generation (UNPG) by combining two PG strategies (i.e., MLPG and CLPG) from a unified perspective to alleviate the mismatch. Moreover, it includes filtering noise-negative pairs, such as too-easy/hard negative pairs, in order to guarantee reliable convergence and improve performance. Consequently, UNPG helps approach WDFS, as shown in Fig. \ref{fig:intro} (d). Through experiments, we demonstrate the superiority of UNPG by achieving state-of-the-art performance using recent loss functions equipped with UNPG on public benchmark datasets  (IJB-B\cite{whitelam2017iarpa}, IJB-C\cite{maze2018iarpa}, MegaFace\cite{kemelmacher2016megaface}, LFW\cite{huang2008labeled}, CFPFP\cite{sengupta2016frontal}, AgeDB-30\cite{moschoglou2017agedb}, CALFW\cite{zheng2017cross}, and K-FACE\cite{choi2021k}) and deliver an in-depth analysis of the reasons behind UNPG. The contributions of this study are summarized as follows:
\begin{itemize}
  \item We propose unified negative pair generation (UNPG), which alleviates the mismatch between the distributions of the sampled pairs and entire negative pairs. UNPG is simple and easy to adapt to existing CL functions with only a minor modification.    
  \item We demonstrate the superiority of UNPG by achieving state-of-the-art performance using recent loss functions equipped with UNPG on public benchmark datasets for face verification and identification.
\end{itemize}
\section{Related Works}
FR is one of the most promising computer-vision tasks. In recent times, the combination of the following three factors has contributed to the rapid growth of this technology: 1) introduction to large-scale face datasets\cite{guo2016ms, yi2014learning, sun2015deeply}, 2) development of effective backbone models\cite{krizhevsky2012imagenet, he2016deep, simonyan2014very, hu2018squeeze, szegedy2015going}, 3) novel loss functions\cite{wang2018cosface, deng2019arcface, meng2021magface}. Among them, loss functions have been actively developed and can be categorized into metric and classification losses.

\noindent\textbf{Metric Loss.} Early direct optimization methods include contrastive loss\cite{chopra2005learning, hadsell2006dimensionality} and triplet loss\cite{schroff2015facenet, hoffer2015deep}, which use the similarity between pairs or triplets in the feature space. They try to make positive samples close and push negative samples far away, but often suffer from slow convergence and poor local optima because they only learn 1:1 relationships in positive and negative pairs. Thus, lifted-structure loss\cite{oh2016deep} and N-pair loss\cite{sohn2016improved} were designed to address this issue. They build massive negative samples and one positive sample based on the same anchor point and learn their relationship simultaneously. Subsequently, other methods have been developed to create more informative pairs. Multi-similarity loss\cite{wang2019multi} classifies existing studies into three types of similarities and devises pair mining and pair weighting methods that satisfy them simultaneously. Tuplet-margin loss\cite{yu2019deep} provides a slack margin to prevent overfitting from hard triplets. Despite these efforts, ML still faces a problem: The negative pairs generated by each iteration cannot represent all identities because FR datasets\cite{guo2016ms, yi2014learning, sun2015deeply} usually have more classes than a mini-batch size.

\noindent\textbf{Classification Loss.} Early indirect optimization methods include softmax loss\cite{cao2018vggface2,deepface-recognition,taigman2014deepface}, which uses the similarity between the deep feature and class weight vectors. Softmax loss has been widely applied in classification problems, but it is not appropriate for FR because testing is done by similarity comparison, not classification. Hence, two methods are being investigated to modify the softmax logit to form a feature space suitable for FR. The first is the normalization of the deep feature and class weight vectors\cite{ranjan2017l2, wang2018cosface, wang2018additive} to reduce the gap between the training and test phases mapped to the cosine similarity space. This is motivated by the interpretation of the feature space of studies such as center-loss\cite{wen2016discriminative}, L-softmax\cite{liu2016large}, and NormFace\cite{wang2017normface}. The second is a margin assignment technique\cite{liu2017sphereface, wang2018cosface, wang2018additive, deng2019arcface, meng2021magface}, which is performed in various ways to enhance intra-class compactness and inter-class dispersion. CosFace\cite{wang2018cosface} and ArcFace\cite{deng2019arcface}, which are typical margin-based loss functions, add external and internal margins to cosine angles, respectively. Recently, MagFace\cite{meng2021magface} introduced a new margin and regularizer technique within several constraints that assumed a positive correlation between magnitude and face quality and ensured convergence. This improved FR performance by creating discriminative features. However, in our interpretation, there is a problem that extremely hard negative similarities in the feature space cannot be expressed by the indirect optimization method alone.

\noindent\textbf{Multi Objective Loss \& Unified Loss.}
Multi-objective loss tried to combine two different losses with a mixture weight at the surface level. MixFace\cite{jung2021mixface} attempted to combine the metric and classification losses (i.e., $\mathcal{L}_{mix}=\mathcal{L}_{arc}+\mathcal{L}_{sn-pair}$) with an analysis of their advantages and disadvantages. However, it is a mixture at the surface level and not a unified loss function. According to the Circle-loss \cite{sun2020circle}, the two existing approaches (i.e., metric and classification losses) can be expressed as a unified loss function. It also adds independent weight factors to deal with ambiguous convergence but is limited in generating pairs (e.g., Circle-loss\cite{sun2020circle} used only MLPG).

\section{Methodology}
\noindent\textbf{Unified Loss.} According to \cite{sun2020circle, sohn2016improved}, the classification and metric losses can be expressed as a unified loss function (i.e., cross-entropy loss). Suppose that $\mathcal{\hat{S}}^p= \{s^p_i|i=1,2,...,K\}$ and $\mathcal{\hat{S}}^n= \{s^n_j|j=1,2,...,L\}$ are the similarity scores for $K$ positive and $L$ negative pairs, respectively. Then, the unified loss function is defined as:

\begin{equation}
\begin{split}
    \mathcal{L}^{uni} &= \frac{1}{K}\sum_{i=1}^{K}\mathcal{L}^{uni}_{i}, \\
    \mathcal{L}^{uni}_{i} &= -\log{\frac{e^{\gamma s^p_i}}{e^{\gamma s^p_i} + \sum_{j=1}^{L}e^{\gamma s^n_j}}} = \log{[1+\sum_{j=1}^{L}e^{\gamma(s^n_j-s^p_i)}]}
\end{split}
\label{eq:unified_loss}
\end{equation}
where $\gamma$ is a positive scale factor.

The only difference between the two losses is the method of computing $\mathcal{\hat{S}}^p$ and $\mathcal{\hat{S}}^n$. We break down this step into PG and SC to clearly explain our proposed method without loss of generality.  

\noindent\textbf{Pair Generation (PG).} In a feature space, let us assume that $\bm{x}_{i}$ and $\bm{x}_{j}$ are $i$-th and $j$-th samples in $N$-sized mini-batch and $y_{i}$ and $y_{j}$ are corresponding target classes, respectively. $\bm{w}_{y_j} $ is a class weight vector of $j$-th class. Then, we generate positive and negative pair sets $\mathcal{P}$ and $\mathcal{N}$ for the metric (Eq. \ref{eq:MLPG}) and classification (Eq. \ref{eq:CLPG}) losses, respectively:

\begin{equation}
\begin{split}
    \mathcal{P}^{ml} &= \{(\bm{x}_{i}, \bm{x}_{j})|y_{i} = y_{j}\} \\
    \mathcal{N}^{ml} &= \{(\bm{x}_{i}, \bm{x}_{j})|{y_{i} \ne y_{j}}\}
\end{split}
\label{eq:MLPG}
\end{equation}

\begin{equation}
\label{eq:CLPG}
\begin{split}
    \mathcal{P}^{cl}_{i} &= ({\bm{x}_{i}, \bm{w}_{y_i}}) \\
    \mathcal{N}^{cl}_{i} &= \{(\bm{x}_{i}, \bm{w}_{j})|j \ne {y}_{i} \}
\end{split}
\end{equation}
In ML, a pair is composed of two samples (e.g., $\bm{x}_{i}$ and $\bm{x}_{j}$) from a mini-batch, and is composed of a sample and a class weight vector (e.g., $\bm{x}_{i}$ and $\bm{w}_{y_i}$) in CL. We denote MLPG and CLPG for PG of the metric and classification losses, respectively.  

\noindent\textbf{Similarity Computation (SC).} We can compute the similarity sets $\mathcal{\hat{S}}^p$ and $\mathcal{\hat{S}}^n$ obtained from PG. The metric and classification losses employ the same similarity method for the same type of pair sets (i.e., positive sets $\mathcal{P}^{ml}$ and $\mathcal{P}^{cl}$ and negative sets $\mathcal{N}^{ml}$ and $\mathcal{N}^{cl}$). Recent research has focused on improving the cosine similarity using a margin. Let us define the angle between two vectors as $\Theta(\bm{a},\bm{b})=\arccos{( \bm{a}^{T}\bm{b}/\lVert \bm{a} \rVert \lVert \bm{b} \rVert )}$. Then,  SN-pair\cite{jung2021mixface} computes $\mathcal{\hat{S}}^p$ and $\mathcal{\hat{S}}^n$ for $\mathcal{P}^{ml}$ and $\mathcal{N}^{ml}$ as: 

\begin{equation}
\begin{split}
    \mathcal{\hat{S}}^p &= \{ \cos{\Theta(\bm{x}_{i}, \bm{x}_{j})}|y_{i} = y_{j} \} \\
    \mathcal{\hat{S}}^n &= \{ \cos{\Theta(\bm{x}_{i}, \bm{x}_{j})}|y_{i} \ne y_{j} \}
\end{split}
\label{eq:SN_pair}
\end{equation}

There is a line\cite{deng2019arcface, meng2021magface, wang2018cosface} of research that employs a margin in cosine similarity. 
In CosFace [33], margin $m$ is placed outside cosine for $\mathcal{\hat{S}}^p$. Thus, $\mathcal{\hat{S}}^p$ and $\mathcal{\hat{S}}^n$ are computed for $\mathcal{P}^{cl}_i$ and $\mathcal{N}^{cl}_i$ as:

\begin{equation}
\begin{split}
    \mathcal{\hat{S}}^p_i &= \{\cos{\Theta(\bm{x}_i, \bm{w}_{y_i} ) + m}\} \\
    \mathcal{\hat{S}}^n_i &= \{ \cos{\Theta(\bm{x}_{i}, \bm{w}_{j} )} | j \ne y_{i} \}
\end{split}
\label{eq:cosface}
\end{equation}
On the other hand, ArcFace\cite{deng2019arcface} places margin $m$ inside cosine: 
\begin{equation}
    \mathcal{\hat{S}}^p_i = \{\cos{(\Theta(\bm{x}_{i}, \bm{w}_{y_{i}})+m)}\}
\label{eq:arcface_sp}
\end{equation}
In other research using margins such as SphereFace\cite{liu2017sphereface} and MagFace\cite{meng2021magface}, $\mathcal{\hat{S}}^p$ and $\mathcal{\hat{S}}^n$ can be derived similarly without loss of generality. 

\begin{figure}[t]
  \includegraphics[width=0.9\columnwidth]{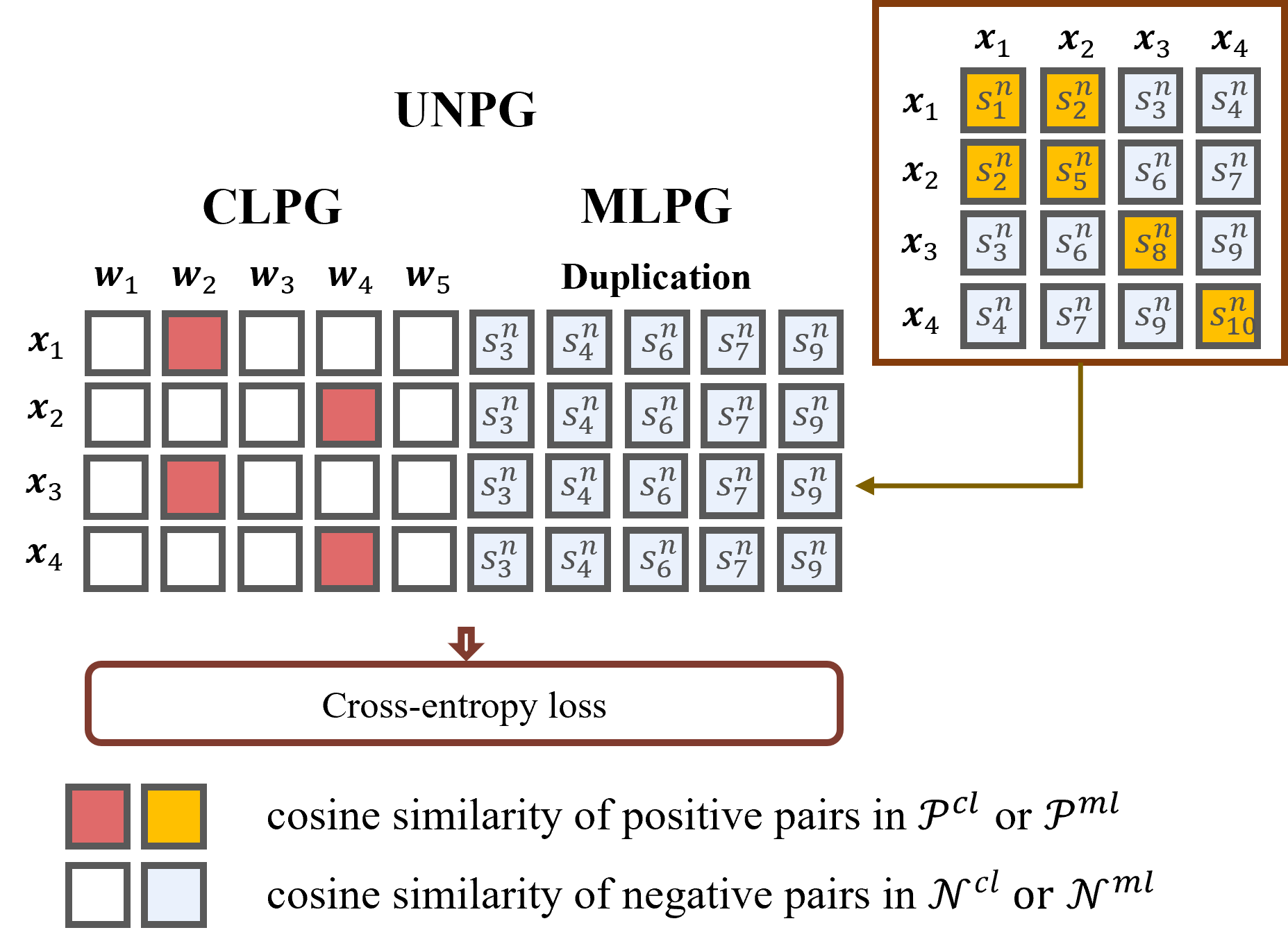}
  \centering
  \caption{
  Unified loss with UNPG.
  }
  \label{uni_vis}
\end{figure}

\noindent\textbf{Unified Negative Pair Generation (UNPG).} We address the fact that PG is the only difference between metric and classification losses from a unified perspective. Previous studies\cite{wang2019multi, yu2019deep, liu2017sphereface, deng2019arcface, wang2018cosface, meng2021magface} attempted to reduce the gap between $\mathcal{S}^p$ and $\mathcal{\hat{S}}^p$ by devising various margin-based methods. Evidently, there is no concern about the gap between $\mathcal{S}^n$ and $\mathcal{\hat{S}}^n$ even though it is a critical component in computing a loss. There are several reasons that cause the gap between $\mathcal{S}^n$ and $\mathcal{\hat{S}}^n$. In ML, it is infeasible to generate negative pairs taking all classes into account in each iteration because of the limited mini-batch size. In CL, it is difficult to generate too-hard negative pairs owing to the convergence of the class weight vectors to their center. This paper proposes unified negative pair generation (UNPG) by combining MLPG and CLPG strategies from a unified perspective to alleviate the mismatches of $(\mathcal{S}^n, \mathcal{\hat{S}}^n)$ and $(\mathcal{S}^p, \mathcal{\hat{S}}^p)$, together. UNPG introduces useful information about negative pairs using MLPG to overcome the CLPG deficiency. In UNPG, a negative pair set $\mathcal{N}^{uni}_{i}$ and the corresponding similarity set $\mathcal{\hat{S}}^n_i$ are defined as:

\begin{equation}
    \begin{gathered}
    \mathcal{N}^{uni}_i = \mathcal{N}^{cl}_i \cup \mathcal{N}^{ml} \\
    \mathcal{\hat{S}}^n_i = \{ \cos{\Theta(\bm{x}_{i}, \bm{w}_{j} )} | j \ne y_{i} \} \cup \{ \cos{\Theta(\bm{x}_{i}, \bm{x}_{j})}|y_{i} \ne y_{j} \}
    \end{gathered}
\label{eq:uni_neg}
\end{equation}
$\mathcal{\hat{S}}^n_i$ can be computed from $\mathcal{N}^{uni}_i$ using various methods such as Eqs. \ref{eq:SN_pair} and \ref{eq:cosface}. Decomposing SC and PG can lead to wide research directions in FR. \noindent As a result, the unified loss with UNPG is defined as:
\begin{equation}
\begin{split}
    \mathcal{L}^{uni}_i = -\log P_i = -\log{\frac{e^{{\gamma s^p_i}}}{e^{{\gamma s^p_i}} + \sum_{j=1}^{L^{cl}}e^{{\gamma s^n_{j}}}+\sum_{k=1}^{L^{ml}} e^{{\gamma s^n_{k}}}}}
\end{split}
\label{eq:unified_loss_revise}
\end{equation}

where $L^{cl}=|\mathcal{N}^{cl}_{i}|$ and $L^{ml}=|\mathcal{N}^{ml}|$.
Note that the normalization term in Eq. \ref{eq:unified_loss_revise} uses the scores from   $\mathcal{N}^{ml}$. Fig. \ref{uni_vis} visualizes Eq. \ref{eq:unified_loss_revise}, where UNPG uses the similarity score matrix obtained from $\mathcal{N}^{ml}$ at each mini-batch and then duplicates it by the size of the mini-batch.


\begin{figure*}[t]
  \centering
  \includegraphics[width=0.9\textwidth]{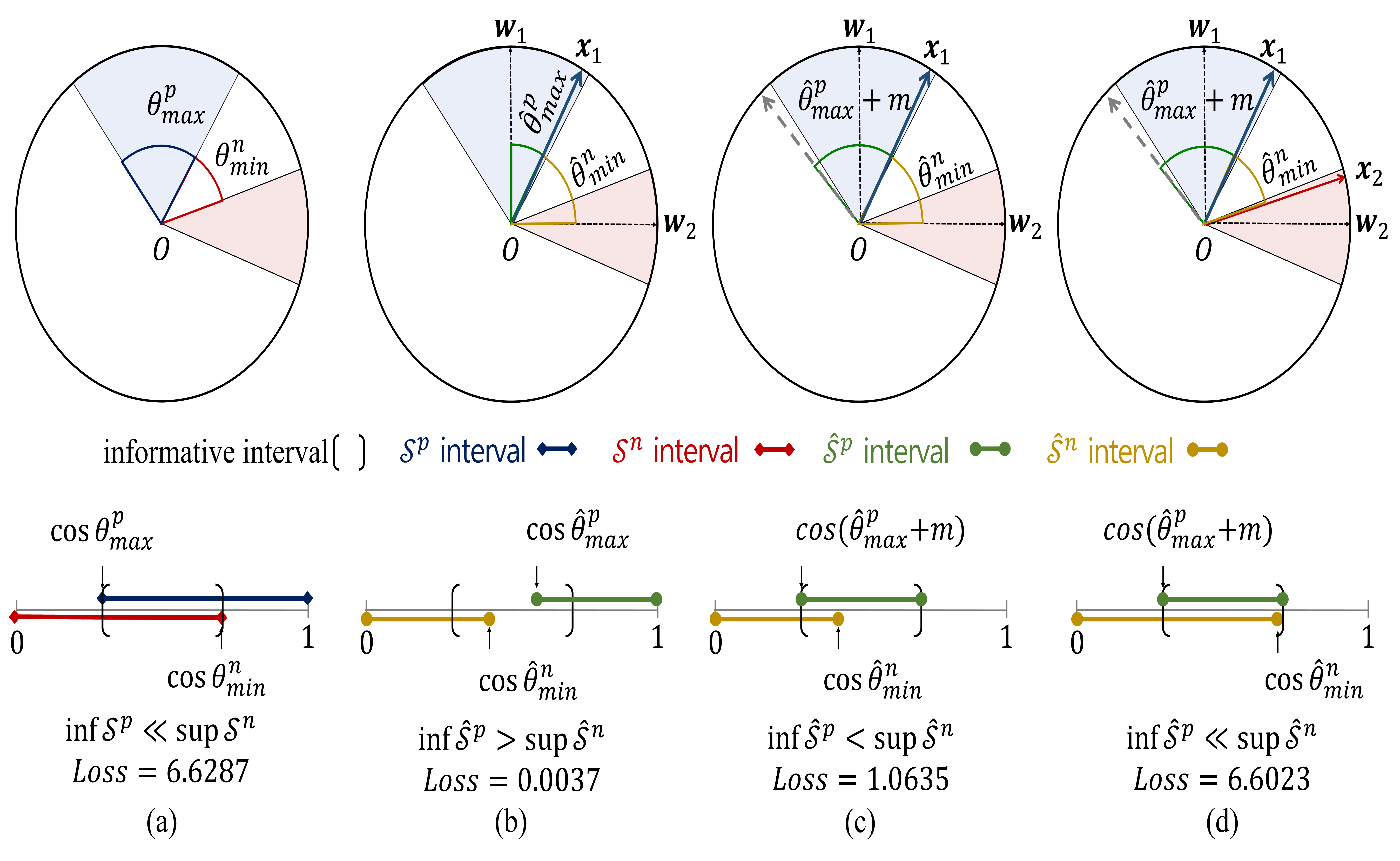}
  \caption{Geometrical interpretation of feature space associated with similarity space. (a) As ideal behavior of the loss function, it imposes a large loss in feature space with low discriminability. A shading area in the same color represents the target region of the same class. $\theta^p_{max}$ and $\theta^n_{min}$are the respective angles of max positive and min negative pairs in the feature space. $\mathcal{S}^p$ and $\mathcal{S}^n$ represent similarity sets. (b) In spite of being equally low discriminative, a very small loss is given by vanilla loss (e.g., norm-softmax). $\bm{w}_1$ and $\bm{w}_2$ are the normalized weight vectors of classes 1 and 2, while $\bm{x}_1$ and $\bm{x}_2$ are the normalized feature vector. $\hat{\theta}^p_{max}$ and $\hat{\theta}^n_{min}$ represent the angle of max positive and min negative pairs in $\hat{\mathcal{S}}^p \subset \mathcal{S}^p$ and $\hat{\mathcal{S}}^n \subset \mathcal{S}^n$, respectively. (c) Mismatch between $\mathcal{S}^p$ and $\mathcal{\hat{S}}^p$ is reduced by using a marinal classification loss (e.g., ArcFace). However, still a small loss is given because of a mismatch between $\mathcal{S}^n$ and $\mathcal{\hat{S}}^n$.  (d) Marginal classification loss with UNPG behaves closest to ideal by alleviating mismatch between $\mathcal{S}^n$ and $\mathcal{\hat{S}}^n$.}
  \label{geo_inter}
\end{figure*}

\begin{algorithm}
\caption{Noise Negative Pair Filtering.}\label{alg:cap}
\begin{algorithmic}
\Require $s^n_j \in \mathcal{\hat{S}}^n$ from $\mathcal{N}^{ml}$, wisker size $r$
\State Extract the lower 25\% similarity $s_l^n$
\State Extract the upper 25\% similarity $s_u^n$
\State IQR $= s^n_u - s^n_l$
\State Min $= s^n_l - r *$IQR, Max $= s^n_u + r *$IQR
\State $\mathcal{\tilde{N}}^{ml} = \{(\bm{x}_i, \bm{x}_j)| y_i \ne y_j \wedge s^n_j >=$Min $\wedge s^n_j <=$Max$\}$
\Ensure $\mathcal{\tilde{N}}^{ml}$
\end{algorithmic}
\end{algorithm}

\noindent\textbf{Noise Negative Pair Filtering.} According to our preliminary experiments, directly utilizing $\mathcal{N}^{ml}$ produced by MLPG causes performance degradation and divergence of a loss because many noise-negative pairs cause a side-effect. We assumed that there are two types of noise pairs: too-easy and too-hard pairs. In the former case, FR models need not pay attention to the pairs but they do, owing to the size of the pairs. In the latter case, FR models cannot allow them because they exceed the representation power of the models. To address this problem, we developed noise-negative pair filtering using a box and whisker algorithm\cite{tukey1977exploratory} as follows:

As a result, UNPG adopting Algorithm 1 is defined as:

\begin{equation}
    \mathcal{N}^{uni}_i = \mathcal{N}^{cl}_i \cup \mathcal{\tilde{N}}^{ml}
\label{eq:uni_neg_prime}
\end{equation}

\noindent\textbf{Geometrical Interpretation of Feature Space} We interpret the role of UNPG by associating a feature space, as shown in Fig. \ref{geo_inter}. To form WDFS satisfying  $\inf{\mathcal{S}^p} > \sup{\mathcal{S}^n}$, a loss function should assign a large loss in the feature space with low discriminability, whereas it should assign a small loss, and vice versa. The unified loss function in Eq. \ref{eq:unified_loss} follows this intent but fails to form WDFS because of the mismatch between similarity sets of the sampled pairs and all pairs. Fig. \ref{geo_inter} (a) depicts the ideal behavior of a loss function that assigns a large loss in the feature space with low discriminability for $\inf{\mathcal{S}^p} \gg \sup{\mathcal{S}^n}$. In contrast, in Fig. \ref{geo_inter} (b), a small loss is assigned in the feature space with low discriminability because sampled $\mathcal{\hat{S}}^p$ and $\mathcal{\hat{S}}^n$ are well-separated. This problem is alleviated using a similarity score with a margin, as shown in Fig. \ref{geo_inter} (c). This makes $\mathcal{\hat{S}}^p$ informative and worthy of training, as the interval of  $\mathcal{\hat{S}}^p$ shifts to the left. Fig. \ref{geo_inter} (d) shows the effect of UNPG. The interval of $\mathcal{\hat{S}}^n$ becomes wider as more negative pairs are included in $\mathcal{\hat{S}}^n$. Consequently, a large loss is assigned, satisfying  $\inf{\mathcal{\hat{S}}^p} > \sup{\mathcal{\hat{S}}^n}$.

\section{Experiments}

\begin{table}[!htbp]
    \caption{A brief overview of FR datasets. (P) and (G) refer to the probe set the gallery set on MegaFace, respectively.}
    \label{tab:datasets}
      \centering
        {\small
        \begin{tabular}[]{ c c c }
         \rowcolor{Cyan!20}
         \hline
         \textbf{Train} & \textbf{\# Identities} & \textbf{\# Images}  \\
         \hline
         MS1M-V2\cite{deng2019arcface} & 85K & 5.8M \\
         K-FACE:T4\cite{choi2021k} & 370 & 3.8M \\
         \hline
         \rowcolor{Cyan!20}
         \textbf{Test} & \textbf{\# Identities} & \textbf{\# Images}  \\
         \hline
         IJB-B\cite{whitelam2017iarpa} & 1,845 & 76.8K \\
         IJB-C\cite{maze2018iarpa} & 3,531 & 148.8K \\
         MegaFace (P)\cite{kemelmacher2016megaface} & 530 & 100K \\
         MegaFace (G)\cite{kemelmacher2016megaface} & 690K & 1M \\
         LFW\cite{huang2008labeled} & 5,749 & 13,233 \\
         CFPFP\cite{sengupta2016frontal} & 500 & 7,000 \\
         AgeDB-30\cite{moschoglou2017agedb} & 568 & 16,488 \\
         CALFW\cite{zheng2017cross} & 5,749 & 12,174 \\
         \rowcolor{Cyan!20}
         \hline
         \textbf{Test}\cite{choi2021k} & \textbf{\# Pairs} & \textbf{\# Variance}  \\
         \hline
         K-FACE:Q1 & 1,000 & Very Low \\
         K-FACE:Q2 & 10K & Low \\
         K-FACE:Q3 & 10K & Middle \\
         K-FACE:Q4 & 10K & High \\
         \hline
        \end{tabular}
        }
\end{table}

\begin{table*}[htb]
    \caption{Verification accuracy of TAR@FAR on IJB-B and IJB-C. “*” indicates results from the original paper.}
    \label{tab:IJBB_IJBC}
      \centering
        {\small
        \begin{tabular}[]{ c c c c c c c c c c c c }
         \hline
         \multirow{2}{*}{\textbf{Method}} & \multirow{2}{*}{\textbf{Backbone}} &  \multicolumn{5}{c}{\textbf{IJB-B(TAR@FAR)}} & \multicolumn{5}{c}{\textbf{IJB-C(TAR@FAR)}} \\
         \cline{3-12}
          & & 1e-6  & 1e-5 & 1e-4 & 1e-3 & 1e-2 & 1e-6 & 1e-5 & 1e-4 & 1e-3 & 1e-2\\
         \hline
         VGGFace2*\cite{cao2018vggface2} & R50 & - & 67.10 & 80.00 & - & - & - & 74.70 & 84.00 & - & - \\
         Circle-loss*\cite{sun2020circle} & R34 & - & - & - & - & - & - & 86.78 & 93.44 & 96.04 & - \\
         Circle-loss*\cite{sun2020circle} & R100 & - & - & - & - & - & - & 89.60 & 93.95 & \textbf{96.29} & - \\
         ArcFace*\cite{deng2019arcface} & R100 & - & - & 94.20 & - & - & - & - & 95.60 & - & - \\
         MagFace*\cite{meng2021magface} & R100 & \textbf{42.32} & \textbf{90.36} & \textbf{94.51} & - & - & \textbf{90.24} & \textbf{94.08} & \textbf{95.97} & - & - \\
         \hline
         Triplet-loss & R34 & 4.42 & 12.57 & 32.65 & 61.33 & 88.78 & 4.04 & 15.32 & 36.86 & 66.46 & 90.77 \\
         contrastive-loss & R34 & 33.10 & 59.40 & 72.18 & 81.98 & 90.11 & 57.84 & 66.41 & 76.16 & 85.03 & 92.21 \\
         CosFace\cite{wang2018cosface} & R34 & 39.70 & 87.47 & 93.55 & 95.71 & 97.05 & 85.95 & 92.57 & 95.23 & 96.81 & 97.94 \\
         \rowcolor{lightgray!20}
         Cos+UNPG & R34 & \textbf{\textcolor{NavyBlue}{43.33}} & 87.51 & 93.58 & \textbf{\textcolor{NavyBlue}{95.96}} & 97.24 & 87.84 & 92.49	& \textbf{\textcolor{NavyBlue}{95.33}} & \textbf{\textcolor{NavyBlue}{96.94}} & \textbf{\textcolor{NavyBlue}{98.06}} \\
         ArcFace & R34 & 40.61 & 86.28 & 93.38 & 95.74 & 97.22 & 85.47 & 92.21 & 95.08 & 96.79 & 97.94 \\
         Arc+Triplet & R34 & 38.31 & 86.46 & 93.22 & 95.72 & 97.28 & 86.40 & 92.19 & 94.97 & 96.68 & 97.94 \\
         Arc+Contrastive & R34 & 38.07 & 86.54 & 93.03 & 95.61 & \textbf{\textcolor{NavyBlue}{97.33}} & 85.21 & 92.54 & 94.86 & 96.60 & 98.01 \\
         \rowcolor{lightgray!20}
         Arc+UNPG & R34 & 40.27	& \textbf{\textcolor{NavyBlue}{88.05}} & \textbf{\textcolor{NavyBlue}{93.66}} & \textbf{\textcolor{NavyBlue}{95.96}} & 97.17 & \textbf{\textcolor{NavyBlue}{87.99}} & \textbf{\textcolor{NavyBlue}{93.02}} & \textbf{\textcolor{NavyBlue}{95.33}} & 96.88 & 97.92 \\
         \hline
         CosFace & R100 & 42.27	& 89.38	& 94.39	& 96.17	& 97.35	& 86.56	& 94.42	& 96.35	& \textbf{\textcolor{OrangeRed}{97.57}}	& 98.26 \\
         \rowcolor{lightgray!20}
         Cos+UNPG & R100 & \textbf{\textcolor{OrangeRed}{49.13}} & 90.61 & 94.99 & 96.50 & 97.36 & 86.95 & 94.48 & \textbf{\textcolor{OrangeRed}{96.39}} & \textbf{\textcolor{OrangeRed}{97.57}} & 98.24 \\
         ArcFace & R100 & 40.68	& 89.99	& 94.89	& 96.40	& 97.59	& 86.57	& 93.93	& 96.25	& 97.43	& 98.31 \\
         \rowcolor{lightgray!20}
         Arc+UNPG & R100 & 44.03 & 90.57 & 95.04 & 96.60 & 97.70 & 88.06 & 94.47 & 96.33 & 97.53 & \textbf{\textcolor{OrangeRed}{98.39}} \\
         MagFace & R100 & 43.71	& 89.03	& 93.99	& 96.11	& 97.32	& 87.19	& 93.30	& 95.54	& 97.00	& 98.05 \\
         \rowcolor{lightgray!20}
         Mag+UNPG & R100 & 46.33 & \textbf{\textcolor{OrangeRed}{90.93}} & \textbf{\textcolor{OrangeRed}{95.21}} & 96.50 & 97.63 & \textbf{\textcolor{OrangeRed}{90.01}} & \textbf{\textcolor{OrangeRed}{94.70}} & 96.38 & 97.51 & 98.32 \\
         \hline
        \end{tabular}
        }
\end{table*}
\subsection{Implementation Details}
\noindent\textbf{Datasets.} For training, MS1M-V2\cite{deng2019arcface} and K-FACE:T4\cite{jung2021mixface} datasets were employed. MS1M-V2, a semi-automatically refined version of MS-Celeb-1M\cite{guo2016ms}, has 5.8M images and 85K identities. K-FACE:T4 is a preprocessed version of K-FACE\cite{choi2021k} utilized in MixFace\cite{jung2021mixface} and has 3.8M images and 370 identities. For testing, several benchmark datasets (IJB-B\cite{whitelam2017iarpa}, IJB-C\cite{maze2018iarpa}, MegaFace\cite{kemelmacher2016megaface}, LFW\cite{huang2008labeled}, CFPFP\cite{sengupta2016frontal}, AgeDB-30\cite{moschoglou2017agedb}, CALFW\cite{zheng2017cross}, and K-FACE:Q1-Q4\cite{jung2021mixface}) were used to evaluate FR models. Table \ref{tab:datasets} summarizes the datasets used in our experiments. 

\noindent\textbf{Training.} For preprocessing, face images were resized to $112\times112$ and normalized using the mean (0.485, 0.456, 0.406) and standard deviations (0.229, 0.224, 0.225). For data augmentation, a horizontal flip was applied with a 50\% of chance. All experiments were performed using two NVIDIA-RTX A6000 GPUs with a mini-batch size of 512. ResNet-34 (R34) and ResNet-100 (R100) were used as backbone models. We re-implemented the state-of-the-art models: CosFace\cite{wang2018cosface}, ArcFace\cite{deng2019arcface}, and MagFace\cite{meng2021magface}. 

\begin{table}[!tb]
    \caption{Identification results on MegaFace datasets with ResNet-100 backbone except for AdaCos. “*” indicates the results from the original paper.}
    \label{tab:megaface}
      \centering
      \begin{center}
        {\small
        \begin{tabular}[]{C{3cm} C{4cm}}
         \hline
         \textbf{Method} & \textbf{Rank-1 accuracy (\%)} \\
         \hline
         AdaCos*\cite{zhang2019adacos} & 97.41  \\
         ArcFace*  & 98.35  \\
         Circle-loss*  & 98.50  \\
         \hline
         MagFace & 98.51 \\
         \rowcolor{lightgray!20}
         Mag+UNPG & 98.03 \\
         ArcFace & 98.56 \\
         \rowcolor{lightgray!20}
         Arc+UNPG & 98.82 \\
         CosFace & 99.08 \\
         \rowcolor{lightgray!20}
         Cos+UNPG & \textbf{\textcolor{OrangeRed}{99.27}} \\
         \hline
        \end{tabular}
        }
      \end{center}
\end{table}

\begin{table}[t]
    \caption{Verification accuracy on LFW, CFP-FP, AgeDB-30, and CALFW with ResNet-100 backbone.}
    \label{tab:lfw}
      \centering
        {\small
        \begin{tabular}[]{c c c c c }
         \hline
         \textbf{Method} & \textbf{LFW} & \textbf{CFP-FP} & \textbf{AgeDB} & \textbf{CALFW} \\
         \hline
         Circle-loss*  & 99.73 & 96.02 & - & -  \\
         ArcFace*  & 99.82 & - & - & 95.45 \\
         MagFace*  & \textbf{99.83} & \textbf{98.46} & \textbf{98.17} & \textbf{96.15} \\
         \hline
         CosFace & \textbf{\textcolor{OrangeRed}{99.83}} & 97.72 & 98.11 & 96.11 \\
         \rowcolor{lightgray!20}
         Cos+UNPG & 99.81 & 98.50 & 98.31 & 96.15 \\
         ArcFace & \textbf{\textcolor{OrangeRed}{99.83}} & 98.60 & 98.23 & 96.11 \\
         \cellcolor{lightgray!20}{Arc+UNPG} & \cellcolor{lightgray!20}{\textbf{\textcolor{OrangeRed}{99.83}}} & \cellcolor{lightgray!20}{98.60} & \cellcolor{lightgray!20}{98.25} & \cellcolor{lightgray!20}{96.13} \\
         MagFace & 99.81 & \textbf{\textcolor{OrangeRed}{98.62}} & 98.30 & 96.15 \\
         \rowcolor{lightgray!20}
         Mag+UNPG & 99.81 & 98.52 & \textbf{\textcolor{OrangeRed}{98.38}} & \textbf{\textcolor{OrangeRed}{96.21}} \\
         \hline
        \end{tabular}
        }
\end{table}

\begin{table*}[!tb]
    \caption{Verification accuracy of TAR@FAR on K-FACE with ResNet-34 backbone.}
    \label{tab:K-FACE}
      \centering
        {\small
        \begin{tabular}[]{ c c c c c c c c c c c c c }
         \hline
         \multirow{2}{*}{\textbf{Method}} & \multicolumn{3}{c}{\textbf{Q4(TAR@FAR)}} & \multicolumn{3}{c}{\textbf{Q3(TAR@FAR)}} & \multicolumn{3}{c}{\textbf{Q2(TAR@FAR)}} & \multicolumn{3}{c}{\textbf{Q1(TAR@FAR)}} \\
         \cline{2-13}
         & 1e-5  & 1e-4 & 1e-3 & 1e-5 & 1e-4 & 1e-3 & 1e-5 & 1e-4 & 1e-3 & 1e-3 & 1e-2 & 1e-1 \\
         \hline
         ArcFace & 0.05 & 0.29 & 4.04 & 2.06 & 4.40 & 18.27 & 26.56 & 41.29 & 63.91 & 94.00 & \textbf{\textcolor{NavyBlue}{100}} & \textbf{\textcolor{NavyBlue}{100}} \\
         AdaCos & 0.28 & 2.57 & 16.68 & 4.11 & 9.94 & 34.57 & 7.16 & 26.31 & 66.88 & 94.00 & \textbf{\textcolor{NavyBlue}{100}} & \textbf{\textcolor{NavyBlue}{100}} \\
         SN-pair\cite{jung2021mixface} & 3.50 & 7.21 & 17.45 & 17.67 & 21.16 & 30.85 & 21.93 & 33.26 & 55.92 & 91.80 & 97.60 & \textbf{\textcolor{NavyBlue}{100}} \\
         MS-loss\cite{wang2019multi} & 5.68	& 8.70 & 25.01 & 15.15 & 18.74 & 38.36 & 38.33 & 46.64 & 66.63 & 94.60 & 99.20 & \textbf{\textcolor{NavyBlue}{100}} \\
         MixFace\cite{jung2021mixface} & 7.11 & 10.92 & 19.92 & 9.19 & 22.55 & 37.67 & 39.09 & 44.48 & 67.44 & 97.00 & \textbf{\textcolor{NavyBlue}{100}} & \textbf{\textcolor{NavyBlue}{100}} \\
         Circle-loss\cite{sun2020circle} &\textbf{\textcolor{NavyBlue}{18.08}} & \textbf{\textcolor{NavyBlue}{25.05}} & \textbf{\textcolor{NavyBlue}{43.46}} & \textbf{\textcolor{NavyBlue}{33.56}} & \textbf{\textcolor{NavyBlue}{41.54}} & \textbf{\textcolor{NavyBlue}{64.88}} & \textbf{\textcolor{NavyBlue}{71.38}} & \textbf{\textcolor{NavyBlue}{77.93}} & \textbf{\textcolor{NavyBlue}{89.97}} & \textbf{\textcolor{NavyBlue}{100}} & \textbf{\textcolor{NavyBlue}{100}} & \textbf{\textcolor{NavyBlue}{100}} \\
         \hline
         \rowcolor{lightgray!20}
         Arc+UNPG &\textbf{\textcolor{OrangeRed}{29.89}} & \textbf{\textcolor{OrangeRed}{50.43}} & \textbf{\textcolor{OrangeRed}{64.05}} & \textbf{\textcolor{OrangeRed}{51.59}} & \textbf{\textcolor{OrangeRed}{60.88}} & \textbf{\textcolor{OrangeRed}{78.68}} & \textbf{\textcolor{OrangeRed}{91.28}} & \textbf{\textcolor{OrangeRed}{93.26}} & \textbf{\textcolor{OrangeRed}{95.68}} & \textbf{\textcolor{OrangeRed}{100}} & \textbf{\textcolor{OrangeRed}{100}} & \textbf{\textcolor{OrangeRed}{100}} \\
         \hline
        \end{tabular}
        }
\end{table*}

The hyper-parameters used in our experiments were as follows: In ArcFace and CosFace, scale factor $\gamma=64$ and margin $m=0.5$ were set. In MagFace, $\gamma=64, l_a = 10, u_a = 110, l_m = 0.4, l_m = 0.8, \lambda_g = 35$ were used. For K-FACE, SN-pair\cite{jung2021mixface} and circle-loss\cite{sun2020circle} employed $\gamma=64$ and $\gamma=32, m=0.25$, respectively. In MixFace\cite{jung2021mixface}, $\epsilon=1e-22$ and $m=0.5$ were set. In MS-loss\cite{wang2019multi}, $\alpha=2,\gamma=0.5,\beta=50$ were used. Triplet loss employed $m = 0.5$. In contrastive loss, positive and negative margins were set to 0 and 1, respectively. Finally, in UNPG, the wisker size $r=1.0$ was used with ResNet-34, whereas $r=1.5$ was used ResNet-100. The stochastic gradient descent (SGD) optimizer was utilized in conjunction with a cosine annealing scheduler\cite{loshchilov2016sgdr} to control the learning rate, which started from 0.1. The momentum, weight decay, and warm-up epochs were set to 0.9, 0.0005, and 3, respectively. The maximum number of training epochs was set to 20 for all models, except that it was set to 25 with MagFace for a fair comparison. The size of the deep feature space extracted from the backbone model was set to 512.

\noindent\textbf{Test.} Cosine similarity was used as a similarity score. Different evaluation metrics were applied depending on the FR tasks. In the verification task (1:1), verification accuracy using the best threshold was exploited for a dataset that has a small number of test images with the same ratio between positive and negative pairs, such as LFW, CFP-FP, AgeDB-30, CALFW, and CPLFW. Otherwise, TAR@FAR was used on IJB-B, IJB-C, and K-FACE. In the identification task (1:N) on MegaFace, rank-1accuracy was utilized.

\begin{figure}[t]
    \centering
    \includegraphics[width=0.9\columnwidth]{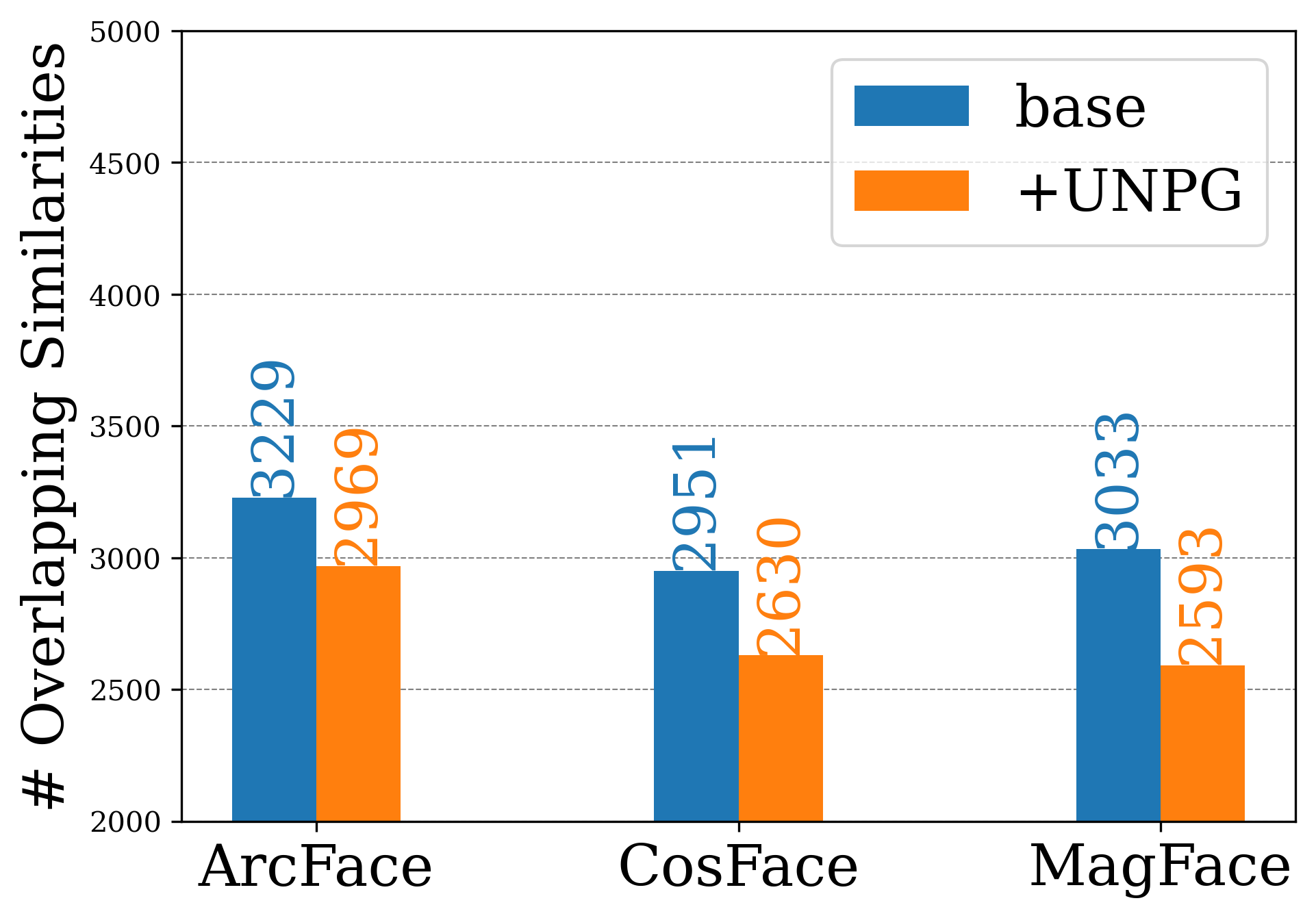}
\caption{Comparison of overlapping similarities for positive and negative pairs with and without UNPG.}
\label{fig:overlap_bar_plot}
\end{figure}

\begin{figure}[t]
    \centering
    \includegraphics[width=0.9\columnwidth]{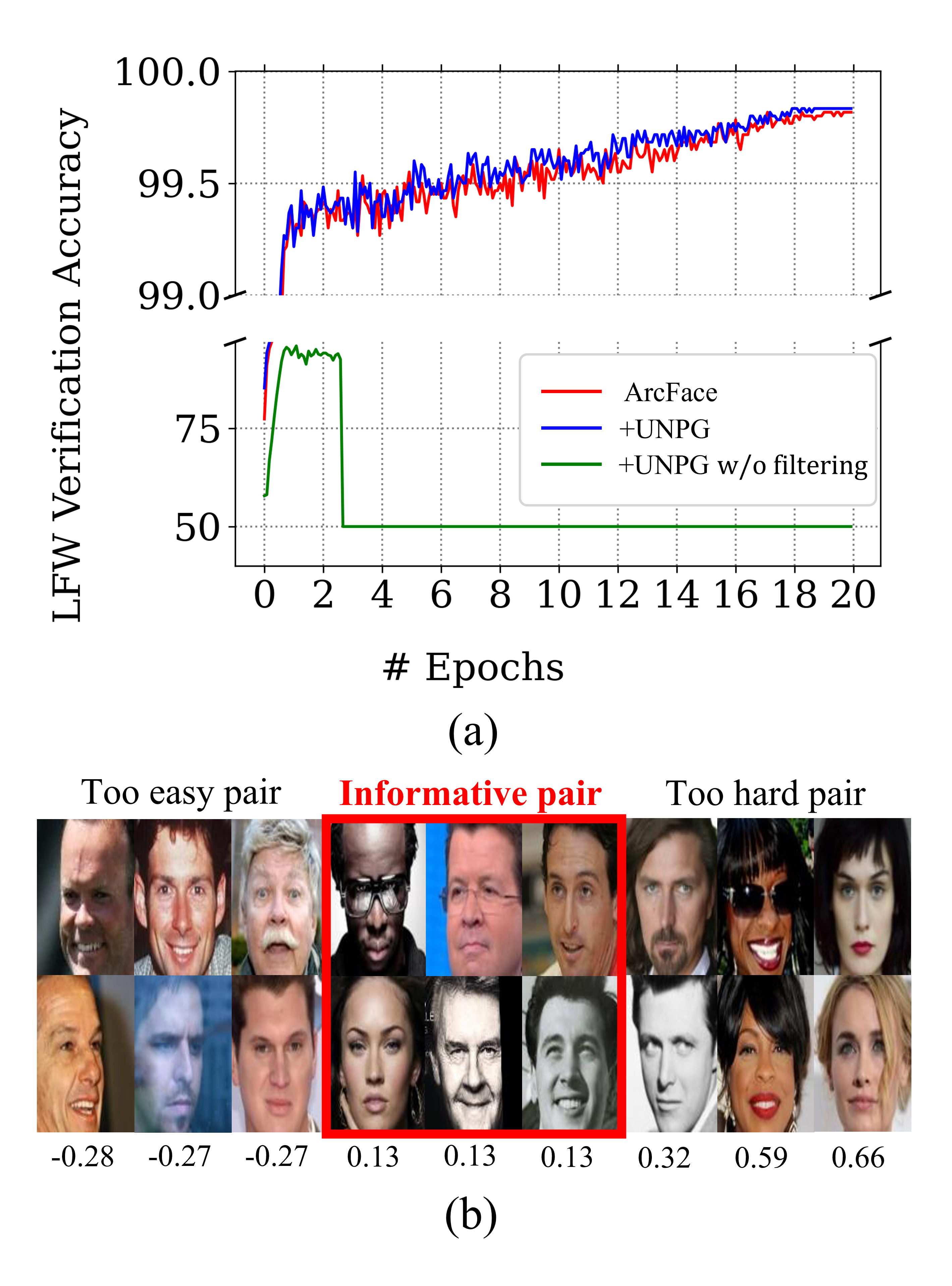}
\caption{(a) Effects of noise negative pair filtering in UNPG with ResNet-34 (b) Examples of three negative pair types in ascending order of similarity scores.}
\label{fig:gap_between_interpretation_and_reality}
\end{figure}

\begin{figure*}[t]
    \centering
    \includegraphics[width=0.8\textwidth]{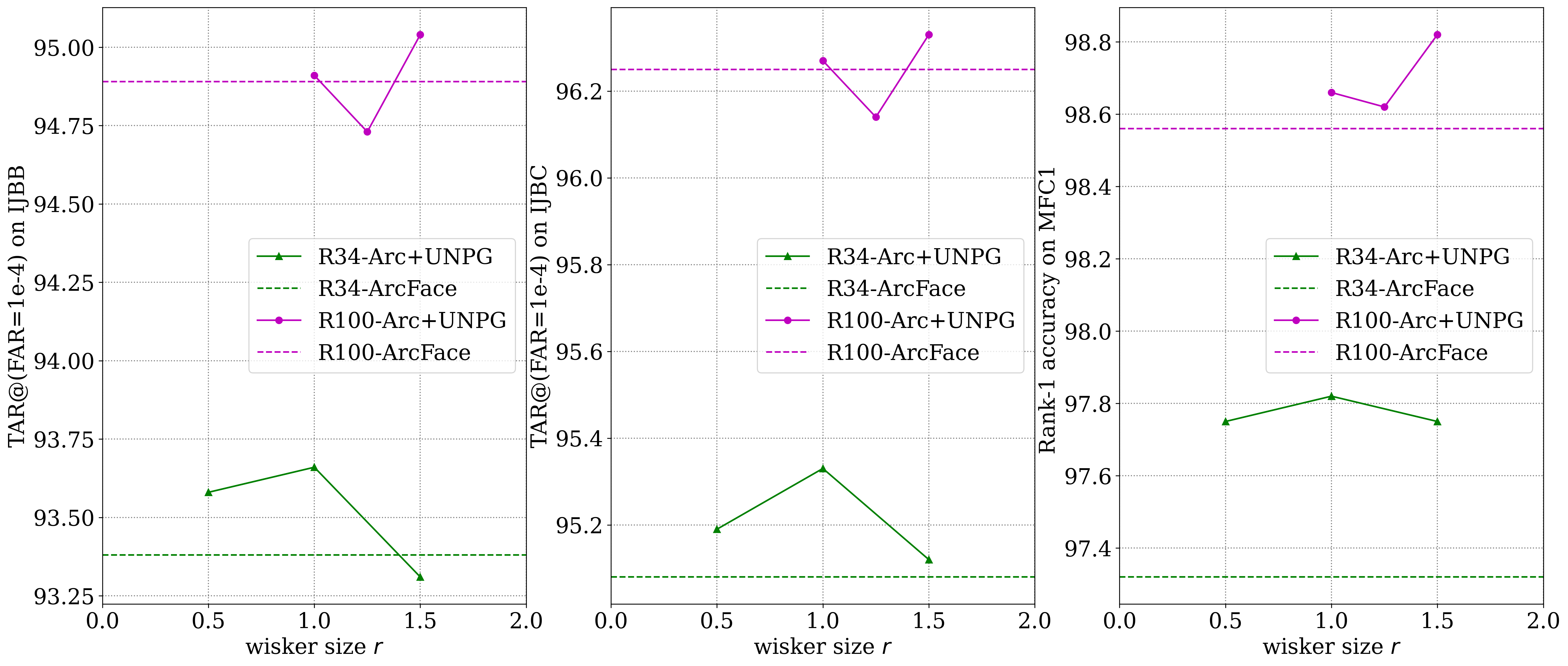}
\caption{Effects of backbone capacity and whisker size on IJB-B (verification), IJB-C (verification), and MegaFace (identification).}
\label{fig:paleto_ijbc_mega}
\end{figure*}

\subsection{Evaluation Results}
\noindent\textbf{Results on IJB-B and IJB-C.} IJB-B consists of 21.8 K images of 1,845 subjects and 55 K frames of 7,011 videos. IJB-C, an extended version of IJB-B, contains 31.3 K images of 3,531 subjects and 117.5 K frames of 11,799 videos. 10 k / 8 M and 19 K / 15 M of positive / negative pairs in IJB-B and IJB-C were used for 1:1 verification. Owing to the severe imbalance between positive and negative pairs, performance was measured by TAR@FAR at different intervals such as [1e-6, 1e-5, 1e-4, 1e-3, 1e-2]. As shown in Table \ref{tab:IJBB_IJBC}, all FR models with UNPG improved at every interval compared to those without UNPG. In particular, TAR@(FAR=1e-4), an interval widely used in FR improved consistently. For example, Mag+UNPG obtained gains of 1.22\% and 0.84\% in IJB-B and IJB-C, respectively, compared to MagFace, and gains of 0.7\% and 0.41\%, respectively, compared to MagFace*.

\noindent\textbf{Results on MegaFace.} MegaFace consists of a gallery set of 1 M images with 690 K classes and probe photos of 100 K images with 530 classes. We followed the test protocol of ArcFace\cite{deng2019arcface}. We removed noisy images and measured rank-1 accuracy for the 1 M distractor after following the identification scenarios using the devkit provided by MegaFace. Table \ref{tab:megaface} present the results of this study. FR models with UNPG performed better than those without it. ArcFace and CosFace using UNPG obtained gains of 0.26\% and 0.19\%, respectively, compared to those without it.

\noindent\textbf{Results on LFW, CFP-FP, AgeDB-30, and CALFW.} FR on LFW, CFP-FP, AgeDB-30, and CALFW is straightforward. Thus, the performance was saturated. LFW, AgeDB-30, and CALFW contain 6,000 images, and CFP-FP has 6,000 images. They have 1:1 ratios between the positive and negative pairs. Verification accuracy was employed with the best threshold separating the positive and negative pairs. In Table \ref{tab:lfw}, the FR models with UNPG obtained competitive performance on the four datasets. 

\noindent\textbf{Results on K-FACE.} K-FACE focuses on FR under fine-grained conditions. It consists of 4.3 M images with 6 accessories, 30 illuminations, 3 expressions, and 20 poses for 400 persons. We adopted the same training and test splits used in MixFace [41]. The training split was composed of 3.8 M images with 370 persons. In particular, the test split, including the remaining 30 persons, was partitioned into Q1, Q2, Q3, and Q4. The number next to Q indicates the variance of conditions where it increases as more conditions are included. Q4 is the most challenging task, whereas Q1 is the most straightforward task among the four. Table \ref{tab:K-FACE} presents the results of the FR models. Surprisingly, ArcFace with UNPG outperformed the other FR models on Q1, Q2, Q3, and Q4. Specifically, it obtained gains of 25.38\%, 19.34\%, and 15.33\% in TAR@(FAR=1e-4) compared to the circle loss.

\subsection{Analysis}

\noindent\textbf{Does it sufficiently satisfy WDFS?} We conclude that UNPG helps FR models to form WDFS by reducing the gap between  $\mathcal{\hat{S}}^p$ and $\mathcal{\hat{S}}^n$. As shown in Fig. \ref{fig:overlap_bar_plot}, we measured the number of overlapping similarity scores between $\mathcal{\hat{S}}^p$ and $\mathcal{\hat{S}}^n$ using ArcFace, CosFace, and MagFace with and without UNPG after training with R100. We randomly sampled 256 positive pairs and 256 most hard negative pairs at each iteration from MS1M-V2. After 1000 iterations, we generated $\mathcal{\hat{S}}^p$ and $\mathcal{\hat{S}}^n$, each with a total of 257,992, and calculated the overlap between them using a histogram. Obviously, applying UNPG reduced the gaps of 260 (ArcFace), 321 (CosFace), and 440 (MagFace) consistently. This proves the effect of UNPG, as expected. 

\noindent\textbf{Effect of Noise-negative Pair Filtering.} To approximate WDFS, $\mathcal{N}^{ml}$ was assumed to include extremely hard negative pairs because it can produce similarity scores similar to $\sup{\mathcal{S}^n}$. In Fig. \ref{fig:gap_between_interpretation_and_reality} (a), we observe that an FR model using $\mathcal{N}^{ml}$ without filtering (+UNPG w/o filtering) at each iteration leads to performance degradation and the divergence of a loss on LFW, whereas it achieves better performance and convergence of a loss with filtering (+UNPG). Although FR models should adequately distinguish too-hard negative pairs ultimately, we argue that it causes adverse effects using a model lacking representation power to cover them. Fig. \ref{fig:gap_between_interpretation_and_reality} (b) shows certain negative pairs, categorized as too-easy, informative, and too-hard pairs, sampled from MS1M-V2, where the similarity scores are computed using ArcFace with UNPG. It is difficult to distinguish two hard pairs, even for humans, because they are very similar. However, too-easy pairs are useless because they are already well separated. Consequently, it is reasonable to focus on the informative pairs in $\mathcal{N}^{ml}$. We can deal with too-hard pairs by enlarging the model capacity, as depicted in Fig. \ref{fig:paleto_ijbc_mega}. We conducted experiments using ArcFace with different backbones, R34 and R100, on IJB-B, IJB-C and MegaFace for verification and identification. In R34, the highest performance was obtained at whisker size $r=1.0$ on all datasets, whereas it was obtained at $r=1.5$ in R100 with a gain of approximately 0.2\%. This reveals that the informative range determined by whisker size $r$ also increases as a model has a large representation power. 

\section{Conclusion}
This paper is based on two insights. First, from a unified perspective, CL and ML have the same purpose of approaching WDFS, except for PG. Second, CL and ML show a mismatch between two similarity distributions of sampled pairs and all negative pairs. Based on these insights, we developed UNPG by combining two PG strategies (MLPG and CLPG) to alleviate the mismatch. Filtering was also applied to remove negative pairs in both too-easy and too-hard pairs. It was observed that UNPG increases the ability to learn existing FR models compared to MLPG and CLPG by providing more informative pairs. Finally, we suggest two research directions in FR: 1) pair generation strategies in the qualitative aspect and 2) loss functions considering the capability of representation power.

\section{Acknowledgement}
This research was supported by Basic Science Research Program through the National Research Foundation of Korea (NRF) funded by the Ministry of Education (NRF-2021R1I1A3052815) and Institute of Information \& communications Technology Planning \& Evaluation (IITP) grant funded by the Korea government (MSIT) (No. 2021-0-00921).

\bibliographystyle{abbrv}
\bibliography{arixv_unpg_cvpr}
\end{document}